\documentclass[sigconf,nonacm]{acmart}
\AtBeginDocument{%
  }

\setcopyright{none}
\usepackage{listings}
\DeclareCaptionStyle{ruled}{labelfont=normalfont,labelsep=colon,strut=off} 
\usepackage{enumitem}
\begin{document}

\title[Perceive, Refine, Reason: Calibrated Visual Measurement]{Perceive, Refine, Reason: A Calibrated Pipeline for Measuring Indicators in Strategic Visual Communication on Social Media}

\author{Weihong Qi}
\email{wq3@iu.edu}
\correspondingauthor
\affiliation{%
  \institution{Indiana University Bloomington}
  \city{Bloomington}
  \state{IN}
  \country{USA}
}

\author{Chen Ling}
\email{ccling@iu.edu}
\affiliation{%
  \institution{Indiana University Bloomington}
  \city{Bloomington}
  \state{IN}
  \country{USA}
}

\renewcommand{\shortauthors}{Qi and Ling.}

\begin{abstract}
Visual content shapes audience perception and opinion on social media, and computational social science increasingly relies on automated tools to analyze images at scale. Yet a measurement gap persists: existing tools rely on predefined categories or produce only coarse image-level labels, while measuring which specific objects appear in an image, how prominently, and where in the frame remains difficult at scale. We introduce \textit{Perceive, Refine, Reason} (PRR), a calibrated pipeline that turns flexible vision-language detectors into auditable measurement instruments for social-scientific research. PRR combines natural-language category prompts with pixel-level spatial refinement via the Segment Anything Model (SAM) and a multimodal LLM arbitration layer whose reasoning chains externalize domain knowledge and lower the expertise threshold for human-in-the-loop validation. A complementary three-tier auditability framework applies quantification learning to profile per-category reliability, support task-aligned configuration, and statistically correct prevalence estimates. Across four vision-language detectors and nine sociological categories, the pipeline yields substantial precision gains over zero-shot baselines, including a 43.3-point improvement for the strongest backbone. Applying PRR to 103{,}920 Facebook images from U.S.\ legislators during the 2024 election cycle and linking detections to DW-NOMINATE ideology scores, we find that more conservative legislators display U.S.\ flags as larger visual elements, with a weaker tendency toward peripheral placement, a spatial pattern invisible to binary detection. PRR provides computational social scientists with a model-agnostic toolkit for accessible, spatially-grounded, and correctable visual measurement.
\end{abstract}




\maketitle

\section{Introduction}
\label{sec:intro}

Visual content on social media often carries communicative intent: the objects that appear in an image, their relative size, and their spatial arrangement can function as rhetorical cues that shape audience perception \citep{schill2012visual, grabe2009image}. To analyze such cues at scale, computational social scientists have developed a growing set of automated tools for visual analysis, including facial recognition and demographic inference \citep{joo2022image, karkkainenfairface2021}, protest detection \citep{won2017protest}, ideology inference \citep{xi2020ideology}, and image clustering \citep{penglu2023, you2017cultural, joshi2024examining}. These tools have enabled researchers to study large-scale visual communication, but they typically rely on predefined category sets or task-specific labeled data and primarily produce categorical annotations, such as image-level labels or detected face and object categories. As a result, many fine-grained and localized visual signals, including object presence, spatial prominence, and co-occurrence patterns, remain difficult to measure at scale without extensive manual coding \citep{penglocksalah2023, torres2022learning}.

Open-vocabulary object detection (OVD) models~\citep{liu2023grounding, minderer2023scaling, cheng2024yoloworld} offer a promising alternative, accepting natural-language prompts and returning localized detections without task-specific training. This flexibility is attractive for computational social science, where researchers seek to operationalize open-ended, theory-driven visual indicators. Yet OVD cannot serve as a social-scientific measurement instrument out of the box. Because these models are trained on natural-image corpora, their predictions misalign with context-dependent sociological constructs: a visually plausible detection may not match the concept a researcher intends to measure. Moreover, bounding-box outputs provide only coarse position and scale, leaving continuous spatial properties such as object area, prominence, and placement within the frame unmeasured.

To address these limitations, we propose \textit{Perceive, Refine, Reason} (PRR), a three-stage calibration pipeline for open-ended visual measurement. The \textit{Perceive} stage uses OVD to generate high-recall candidate detections from natural-language prompts. The \textit{Refine} stage applies the Segment Anything Model (SAM) \citep{kirillov2023segment} to convert bounding boxes into pixel-level masks, enabling continuous measurements of object area, prominence, and location. The \textit{Reason} stage uses a multimodal LLM as an arbitration layer~\citep{chen2024mllmjudge} to determine whether each candidate detection matches the intended sociological construct, producing auditable reasoning chains for semantic validation. This staged decomposition is motivated by two considerations regarding end-to-end multimodal LLM inference in our setting. First, the spatial coordinates produced by current multimodal LLMs are typically less precise than those of specialized detectors~\citep{vanguard2026, xiao2026visualgrounding}, which limits their suitability for the pixel-level measurements our pipeline depends on. Second, end-to-end LLM detection requires complex multi-object reasoning per call, escalating cost and latency at corpus scale. Beyond instance-level validation, we further introduce a three-tier auditability framework, grounded in quantification learning \citep{forman2008quantifying} and statistically principled measurement from web data \citep{lenti2024likelihood}, that enables researchers to profile category-specific reliability, select task-appropriate configurations, and correct aggregate frequency estimates.

This work contributes to computational social science by reframing open-vocabulary vision models as auditable measurement instruments for social media imagery. Specifically, we make three contributions:
\begin{enumerate}[nosep,leftmargin=*]
    \item \textbf{A calibrated pipeline for open-ended visual measurement}. We introduce PRR, a model-agnostic pipeline that combines off-the-shelf OVD, SAM-based pixel-level spatial quantification, and LLM-based semantic arbitration. The pipeline enables researchers to move beyond image-level labels and bounding boxes toward construct-aware and spatially grounded visual measurements.
    
    \item \textbf{An auditability framework for measurement reliability}. We develop a three-tier protocol that profiles category-specific reliability, supports task-aligned model configuration, and statistically corrects aggregate prevalence estimates. This framework makes detection errors explicit and correctable rather than allowing them to silently propagate into downstream social-scientific findings.
    
    \item \textbf{A systematic evaluation and political communication case study}. We evaluate four OVD architectures across nine sociological categories and apply PRR to 103{,}920 Facebook images posted by U.S.\ legislators during the 2024 election cycle. The case study shows that ideological differences in visual communication appear not only in whether legislators display U.S.\ flags, but also in how prominently and where in the frame those flags are placed, patterns that categorical detection alone cannot capture.
\end{enumerate}

Together, PRR and the auditability framework provide a practical toolkit for defining visual indicators in natural language, extracting spatially grounded measurements from large image collections, auditing reliability boundaries, and conducting downstream analyses with documented and statistically correctable measurement error.

\section{Related Work}
\subsection{Visual Analysis in Computational Social Science}
A growing body of research applies computer vision to social media imagery for substantive social-scientific questions. Studies have classified protest activity~\citep{won2017protest}, inferred political ideology from legislators' visual self-presentation~\citep{xi2020ideology}, examined state-sponsored image coordination~\citep{ng2022coordinated}, identified ideologically correlated imagery in political communication~\citep{joshi2024examining}, and clustered photographs to track cultural and thematic patterns over time~\citep{penglu2023, you2017cultural}. Adjacent methodological work supports these efforts through facial recognition and demographic inference~\citep{joo2022image, karkkainenfairface2021}, scene classification~\citep{zhou2018places}, and tutorials on adapting convolutional architectures for social-science research~\citep{torres2022learning}. However, two methodological constraints recur across this literature. First, existing tools operate on predefined category sets or require labeled training data, restricting their flexibility whenever a new sociological indicator becomes of interest. Second, they primarily produce categorical annotations: a single label assigned to a whole image or to detected faces and objects, leaving fine-grained spatial properties such as object size, centrality, and co-occurrence unmeasurable at scale~\citep{penglocksalah2023, grabe2009image, schill2012visual}.

\subsection{Open-Vocabulary Object Detection}
Open-vocabulary object detection (OVD) directly addresses the closed-category limitation of conventional vision-based tools. OVD models accept natural-language prompts as category definitions and return localized detections without task-specific training. Representative architectures include GroundingDINO~\citep{liu2023grounding}, which combines DINO with grounded language pre-training, the OWL-ViT family~\citep{minderer2022simple, minderer2023scaling}, which adapts CLIP-pretrained Vision Transformers for detection, and YOLO-World~\citep{cheng2024yoloworld}, which augments a CNN backbone with offline vocabulary encoding for real-time inference. These OVD architectures lift the closed-vocabulary restriction and recover spatial localization, but two challenges limit their direct application to social-scientific measurement. First, OVD models are pre-trained on natural-object corpora, producing systematic mismatches with sociological constructs that are context-dependent and visually ambiguous. Second, bounding-box outputs encode only coarse position and scale rather than pixel-level extent, restricting the granularity of downstream spatial measurements.

\subsection{Bridging Vision Models and Social Science Measurement}
Addressing the limitations of OVD requires combining ideas from several adjacent literatures. Quantification learning~\citep{forman2008quantifying} establishes that biased classifiers yield unbiased prevalence estimates under documented Precision and Recall profiles, an insight extended to text-based estimation~\citep{hopkins2010method} and echoed in web mining's turn toward statistically principled measurement and deployment-realistic evaluation~\citep{lenti2024likelihood, kim2025revisiting}, but rarely applied to visual analysis. LLM-based annotation matches or exceeds human annotators on social media labeling tasks while producing natural-language justifications that support verification~\citep{ziems2024can, gilardi2023chatgpt}, and multimodal LLMs (MLLMs) have been explored as standalone analyzers of social-media images~\citep{lyu2023gpt4v, mina2026}. Yet MLLMs alone exhibit persistent hallucinations, inconsistent judgments, and spatial coordinates less precise than specialized detectors~\citep{chen2024mllmjudge, vanguard2026, xiao2026visualgrounding}, motivating designs that combine MLLM semantic reasoning with detector-based spatial grounding; the Segment Anything Model~\citep{kirillov2023segment} adds pixel-level measurement to box-based detectors. Our work brings these lines together: the pipeline is model-agnostic and paired with a three-tier audit protocol that documents and statistically corrects detection error.


\section{Method}
\label{sec:method}

\subsection{Pipeline Overview}
\label{sec:overview}

\begin{figure*}[!htbp]
  \centering
  \includegraphics[width= .95\textwidth]{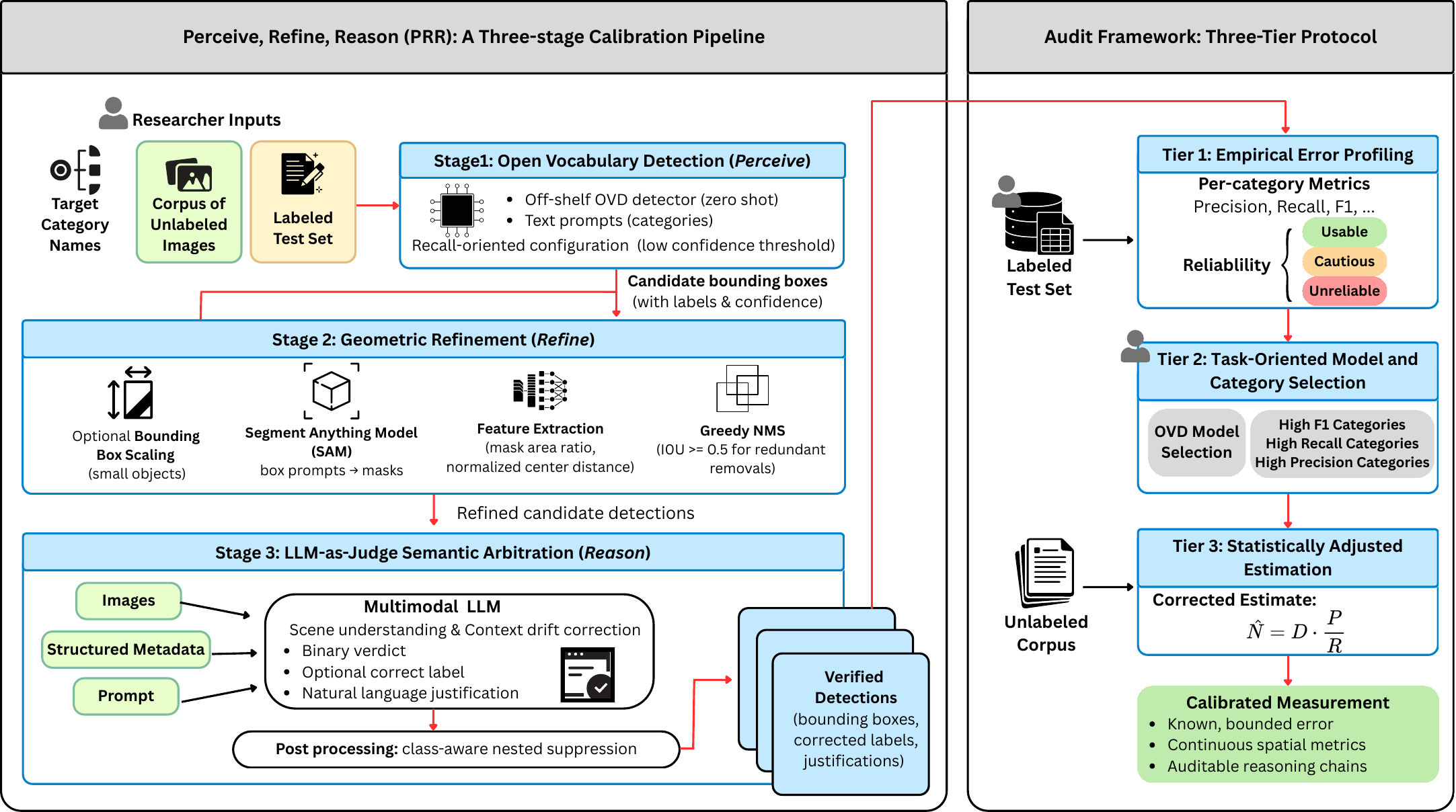}
 \caption{Overview of the PRR pipeline (left) and the three-tier audit framework (right). The pipeline transforms raw open-vocabulary detections into verified detections with spatial measurements and auditable reasoning. The audit framework profiles per-category reliability on a labeled test set (Tier 1), guides task-aligned configuration (Tier 2), and applies statistical correction to prevalence estimates on the full corpus (Tier 3), yielding calibrated measurements with known, bounded error. The human icon marks stages requiring researcher input or decisions; other stages are automated.}
  \label{fig:flow_chart}
  \vspace{-12pt}
\end{figure*}

We propose a model-agnostic calibration pipeline that transforms raw open-vocabulary detections into research-quality annotations through three sequential stages: \textit{Perceive}, \textit{Refine}, and \textit{Reason}. It accepts any off-the-shelf OVD detector as its perception front-end and applies a shared SAM and LLM calibration backbone, bridging the semantic gap without architecture-specific modifications. To deploy the pipeline, researchers provide (1)~a set of target category names, (2)~a corpus of unlabeled images, and (3)~a labeled test set for auditing. The pipeline returns, for each image, a set of verified detections that include bounding boxes, pixel-level masks, corrected labels, and natural-language justifications, supporting both transparent measurement and qualitative auditability of the automated annotation process. Figure~\ref{fig:flow_chart} summarizes the pipeline with the three-tier audit framework.

\begin{figure*}[!htbp]
  \centering
  \includegraphics[width= .95\textwidth]{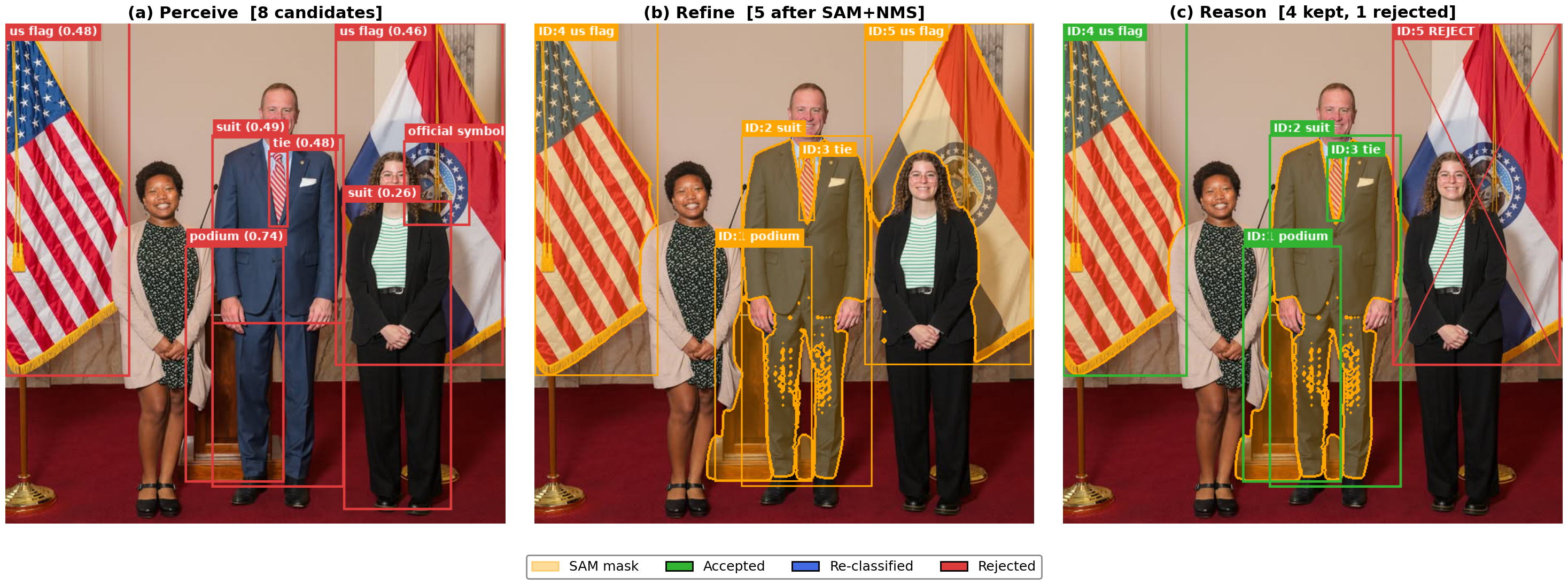}
    \caption{Overview of the PRR pipeline applied to a single image posted on Facebook by U.S. senator Eric Schmitt. \textbf{(a) Perceive:} An Open-Vocabulary Detector (OVD) generates 8 raw candidate boxes with noisy labels and redundancies. \textbf{(b) Refine:} The Segment Anything Model (SAM) extracts pixel-level masks (orange) for spatial grounding, and Non-Maximum Suppression (NMS) reduces the candidate set to 5. \textbf{(c) Reason:} An LLM evaluates this refined set to accept (green), reject (red), or re-classify (blue) each candidate.}
  \label{fig:single_image_pipeline}
  \vspace{-12pt}
\end{figure*}

\subsection{Stage 1: Open-Vocabulary Detection (Perceive)}
\label{sec:perceive}

The \textit{Perceive} stage deploys an off-the-shelf OVD detector in a zero-shot setting. Target categories provided by the researcher are passed as text prompts to the model, which returns candidate bounding boxes, associated labels, and confidence scores without task-specific fine-tuning. We adopt low confidence thresholds to maximize recall, ensuring broad capture of potential visual indicators and delegating false-positive filtering to the downstream stages. The stage is agnostic to the choice of detector: any vision-language model that processes text-conditioned queries and outputs labeled bounding boxes can serve as the perception front-end.

\subsection{Stage 2: Geometric Refinement (Refine)}

Raw OVD outputs are axis-aligned rectangles that carry no information about object shape. To obtain finer-grained geometry, candidate boxes are passed to the Segment Anything Model (SAM; ViT-H)~\citep{kirillov2023segment} as box prompts, producing pixel-level masks. For small-object categories, bounding boxes are optionally scaled prior to SAM processing to prevent mask over-expansion. A greedy non-maximum suppression (NMS) step~\citep{neubeck2006efficient, felzenszwalb2010object} then removes redundant detections (IoU $\geq 0.5$). NMS is applied after SAM refinement rather than directly after OVD inference: SAM-refined boxes yield tighter IoU computations, supporting more precise deduplication and reducing the candidate set before the more costly LLM evaluation. The resulting pixel-level masks support extraction of continuous spatial features such as mask area ratio and normalized center distance, which serve as quantitative measures of object prominence and inform downstream analyses, for example by distinguishing a small background flag from one that dominates the frame.

\subsection{Stage 3: Semantic Arbitration via LLM-as-Judge (Reason)}
\label{sec:reason}

OVD detectors handle objects with stable visual prototypes well but struggle with context-dependent sociological constructs, producing three failure modes: near-zero recall for abstract concepts that lack a fixed visual prototype; \textit{context drift}, where detectors localize valid objects but assign semantically adjacent yet incorrect labels; and semantic false positives produced by the recall-oriented thresholds of the \textit{Perceive} stage. The \textit{Reason} stage addresses these failures with a multimodal LLM arbitration layer. Refined candidate regions, overlaid with numeric ID labels, are submitted to the LLM alongside structured metadata describing each candidate; the prompt embeds domain-specific heuristics and a strict category whitelist, and instructs the LLM to \textit{re-classify} visually valid but mislabeled detections. For each candidate, the model returns a binary accept/reject verdict, a corrected label when applicable, and a natural-language reasoning chain. Figure~\ref{fig:single_image_pipeline} illustrates the three stages on a single example: an OVD pass yields 8 candidate boxes, SAM and NMS reduce them to 5 refined detections, and the LLM accepts 4 and rejects 1. A class-aware nested suppression step then removes redundant same-class detections (intersection-over-minimum area $> 70\%$), while cross-class nesting is preserved to retain sociologically meaningful co-occurrences such as a \textit{tie} within a \textit{suit}.


\section{Auditability Framework}
\label{sec:audit}

No automated vision pipeline achieves perfect accuracy. For valid scientific inference, what matters is whether the residual errors are \textit{known}, \textit{quantifiable}, and \textit{correctable}. Drawing on quantification learning~\citep{forman2008quantifying}, which formalizes how aggregate class prevalence can be estimated from imperfect classifiers, we introduce a three-tier audit protocol that is applied to a labeled test set before the pipeline is deployed on an unlabeled corpus, transforming an imperfect detector into a calibrated measurement instrument with documented bias bounds. This shifts the methodological standard for vision-based social-science measurement from \textit{detector accuracy} to \textit{inference validity}~\citep{kim2025revisiting}: a pipeline need not be uniformly accurate to support valid claims, provided its errors are characterized and the analyses are scoped to what those errors permit.

\paragraph{Tier 1: Empirical Error Profiling.}
The pipeline is first evaluated on a labeled test set drawn from the target domain, yielding per-category Precision ($P$) and Recall ($R$). The \textit{acceptability} of these metrics is determined not by universal computer vision benchmarks but by the requirements of the downstream analytical task. A category with $F1 = 55\%$ and $P = 93\%$, for instance, may be adequate for comparative group analyses that prioritize low measurement noise, yet inadequate for absolute frequency estimation. This empirical profile grounds the methodological decisions that follow.

\paragraph{Tier 2: Task-Aligned Configuration.}
The empirical profile from Tier 1 informs configuration choices that depend on the analytical task. For instance, hypothesis testing and regression analyses are sensitive to false positives, which introduce systematic measurement noise; these tasks call for high-precision configurations. Exploratory content analysis and frequency estimation are sensitive to false negatives, which undercount symbol prevalence; these tasks favor high-recall configurations. Tier 2 selects, for each category, the model and threshold that best align the empirical $P$--$R$ profile with the inferential goals of the study.

\paragraph{Tier 3: Statistical Prevalence Calibration.}
Once a configuration is fixed, Tier 3 corrects aggregate prevalence estimates for the full corpus. Let $D$ denote the raw detection count for a given category. The corrected prevalence estimate $\hat{N}$ is computed as:
\begin{equation}
\hat{N} = D \cdot \frac{P}{R}
\label{eq:tier3}
\end{equation}
Multiplying by $P$ discounts expected false positives, while dividing by $R$ compensates for undetected true instances. Under the standard supervised-learning assumption that test-set error rates generalize to the target corpus, this estimator corrects for the documented measurement bias profiled in Tier 1. Confidence intervals are obtained by bootstrap resampling of the test-set predictions.

\section{Case Study: Visual Indicators in U.S.\ Legislative Communication}
\label{sec:case_study}

We apply the pipeline and auditability framework to a large-scale corpus of social media imagery published by U.S.\ legislators. Localized visual indicators in this domain, including national flags, formal attire, and public speaking arrangements, carry deliberate rhetorical weight, yet have eluded reliable quantification at scale under existing computer vision tools.

\subsection{Data}
\label{sec:data}

We identified the official Facebook accounts of all 540 members of the 118th U.S.\ Congress, locating 472 active profiles (232 Democrats, 237 Republicans, 3 Independents). From these accounts we collected 103,920 images posted between June 15 and November 15, 2024, a window spanning the final five months of the 2024 election cycle. Posting activity is balanced across both party affiliation and chamber, with comparable mean and median image counts per legislator across groups (Table~\ref{tab:summary_stats}).

We merged the image corpus with DW-NOMINATE scores~\citep{poole1985spatial} from the Voteview project for both chambers of the 118th Congress. A multi-stage name-matching procedure that handles nicknames, title variations, and special characters linked 458 of 472 accounts (97.0\%) to their dimension 1 and dimension 2 scores. Unmatched accounts, primarily non-voting delegates and staff-managed pages, are retained in the corpus but excluded from ideology-based analyses.

For evaluation, we drew a random sample of 105 images and obtained exhaustive bounding-box annotations from two trained coders across nine target categories. Disagreements were resolved through qualitative reconciliation to produce a consensus ground truth.

\begin{table}[h!]
\centering
\resizebox{\columnwidth}{!}{
\begin{tabular}{lccc}
\toprule
\textbf{Group} & \textbf{Members} & \textbf{Mean Images} & \textbf{Median Images} \\
\midrule
Democrats (D) & 232 & 216.49 & 200.0 \\
Independents (I) & 3 & 170.00 & 143.5 \\
Republicans (R) & 237 &  214.40 & 179.0 \\
\midrule
House of Representatives & 379 & 220.59 & 189.0 \\
Senate & 93 & 192.95 & 187.0 \\
\bottomrule
\end{tabular}
}
\caption{Summary statistics of collected images by party affiliation and congressional chamber.}
\label{tab:summary_stats}
\vspace{-10pt}
\end{table}

\subsection{Target Categories}
\label{sec:categories}

We define nine target categories drawn from the visual political communication literature~\citep{schill2012visual, grabe2009image}: \textit{U.S.\ flag}, \textit{tie}, \textit{suit}, \textit{podium}, \textit{solo speaker}, \textit{official symbol}, \textit{military}, \textit{organization symbol}, and \textit{politician symbol}. The set is constructed to span a spectrum of semantic and visual abstraction. At the concrete end, \textit{U.S.\ flag}, \textit{tie}, and \textit{podium} have stable visual prototypes that conventional OVD models recognize reliably. \textit{Suit} and \textit{solo speaker} introduce contextual ambiguity that requires relational reasoning rather than pixel matching: distinguishing a formal blazer from casual wear, or determining whether a person is anchored to an audience and a podium. At the abstract end, \textit{official symbol}, \textit{organization symbol}, and \textit{politician symbol} lack any fixed morphological form. This gradient lets us diagnose the boundary conditions of zero-shot OVD and quantify the marginal value of LLM-driven semantic calibration across levels of abstraction.

\subsection{Human Annotation and Ground Truth}
\label{sec:human_annotation}

We constructed the ground truth in three steps. First, we developed an annotation codebook with operational rules covering tight enclosure, single-instance correspondence, complete visible extent, exclusion of ambiguous shapes, and separate guidelines for bounding-box and polygon-mask annotation. Second, two annotators trained on the codebook independently labeled all 105 sampled images with both bounding boxes and pixel-level polygon masks. Third, the annotators reconciled disagreements through targeted discussion to produce the final consensus annotations.


Inter-coder reliability prior to reconciliation was high: the macro-average Intersection over Union (IoU) was 0.82 for bounding boxes and 0.85 for polygon masks, substantially above the IoU $\geq 0.5$ threshold used in object detection benchmarks~\citep{everingham2010pascal, lin2014coco} and comparable to Cityscapes~\citep{cordts2016cityscapes}. The 105 images yield 220 annotated object instances across the nine categories. Because the test set is used to estimate error rates rather than to train models, its size enters the analysis as quantified uncertainty: bootstrap confidence intervals are reported for all audit-derived quantities, and analyses are scoped to what those intervals support.


 
\subsection{Experimental Setup}
\label{sec:models}
\paragraph{OVD Baselines.}
To demonstrate the model-agnostic nature of our calibration pipeline, we evaluate it across four off-the-shelf OVD detectors spanning distinct architectural paradigms. \textbf{Grounding DINO} \citep{liu2023grounding} represents state-of-the-art deep cross-modal fusion. \textbf{OWL-ViT~v1} \citep{minderer2022simple} applies contrastive alignment to a CLIP-pretrained Vision Transformer, and its self-trained successor \textbf{OWL-v2} \citep{minderer2023scaling} allows us to assess generational architectural improvements. \textbf{YOLO-World} \citep{cheng2024yoloworld} augments a CNN-based architecture with offline vocabulary encoding, offering a real-time, efficiency-oriented paradigm. All models are deployed in a strict zero-shot setting using only the nine target category names as prompts, without task-specific fine-tuning or domain adaptation.

\paragraph{Implementation Details.}
At the \textit{Perceive} stage, confidence thresholds are set conservatively low to maximize initial recall, delegating precision filtering to downstream stages. Specific thresholds vary across models to accommodate differences in internal score calibration: $\tau = 0.2$ for OWL-v2 and OWL-ViT~v1, $\tau = 0.05$ for YOLO-World, and $\tau_{\text{box}} = 0.35$ with $\tau_{\text{text}} = 0.25$ for Grounding DINO. The \textit{Refine} stage uses the default Segment Anything Model (SAM, ViT-H checkpoint) \citep{kirillov2023segment} for mask generation, followed by non-maximum suppression with an IoU threshold of $0.5$. The \textit{Reason} stage employs \texttt{gpt-4.1-mini} \citep{openai2025gpt41} as the multimodal arbitrator, with generation temperature fixed at $0.0$ to ensure deterministic outputs. Class-aware nested suppression uses an intersection-over-minimum area threshold of $0.7$. All local inference, including OVD and SAM operations, was executed on a single NVIDIA A100 80GB PCIe GPU.

\section{Experimental Results}

\begin{table}[!htbp]
\centering
\small
\setlength{\tabcolsep}{6pt}
\begin{tabular}{llccc}
\toprule
Model & Variant & Precision & Recall & F1$\uparrow$ \\
\midrule
GroundingDINO
& Zero-shot  & 16.6 & \textbf{25.1} & 20.0 \\
& Calibrated & \textbf{45.1} & 21.2 & \textbf{28.9} \\
&            & {\color{olive}(+28.5)} & {\color{olive}(-3.9)} & {\color{olive}(+8.9)} \\
\midrule
OWL-ViT v1
& Zero-shot  & 66.3 & \textbf{30.7} & \textbf{42.0} \\
& Calibrated & \textbf{92.0} & 17.6 & 29.6 \\
&            & {\color{olive}(+25.7)} & {\color{olive}(-13.1)} & {\color{olive}(-12.4)} \\
\midrule
OWL-v2
& Zero-shot  & 36.9 & \textbf{50.1} & 42.5 \\
& Calibrated & \textbf{80.2} & 33.2 & \textbf{47.0} \\
&            & {\color{olive}(+43.3)} & {\color{olive}(-16.9)} & {\color{olive}(+4.5)} \\
\midrule
YOLO-World
& Zero-shot  & 36.9 & \textbf{21.0} & 26.8 \\
& Calibrated & \textbf{65.5} & 18.4 & \textbf{28.7} \\
&            & {\color{olive}(+28.6)} & {\color{olive}(-2.6)} & {\color{olive}(+1.9)} \\
\bottomrule
\end{tabular}
\caption{Overall micro-averaged performance of zero-shot and calibrated variants. Best results for each model are highlighted in bold. Numbers in color indicate absolute changes of the calibrated variant over the zero-shot baseline.}
\label{tab:main_results}
\end{table}

\begin{figure*}[!t]
  \centering
  \includegraphics[width= .9\textwidth]{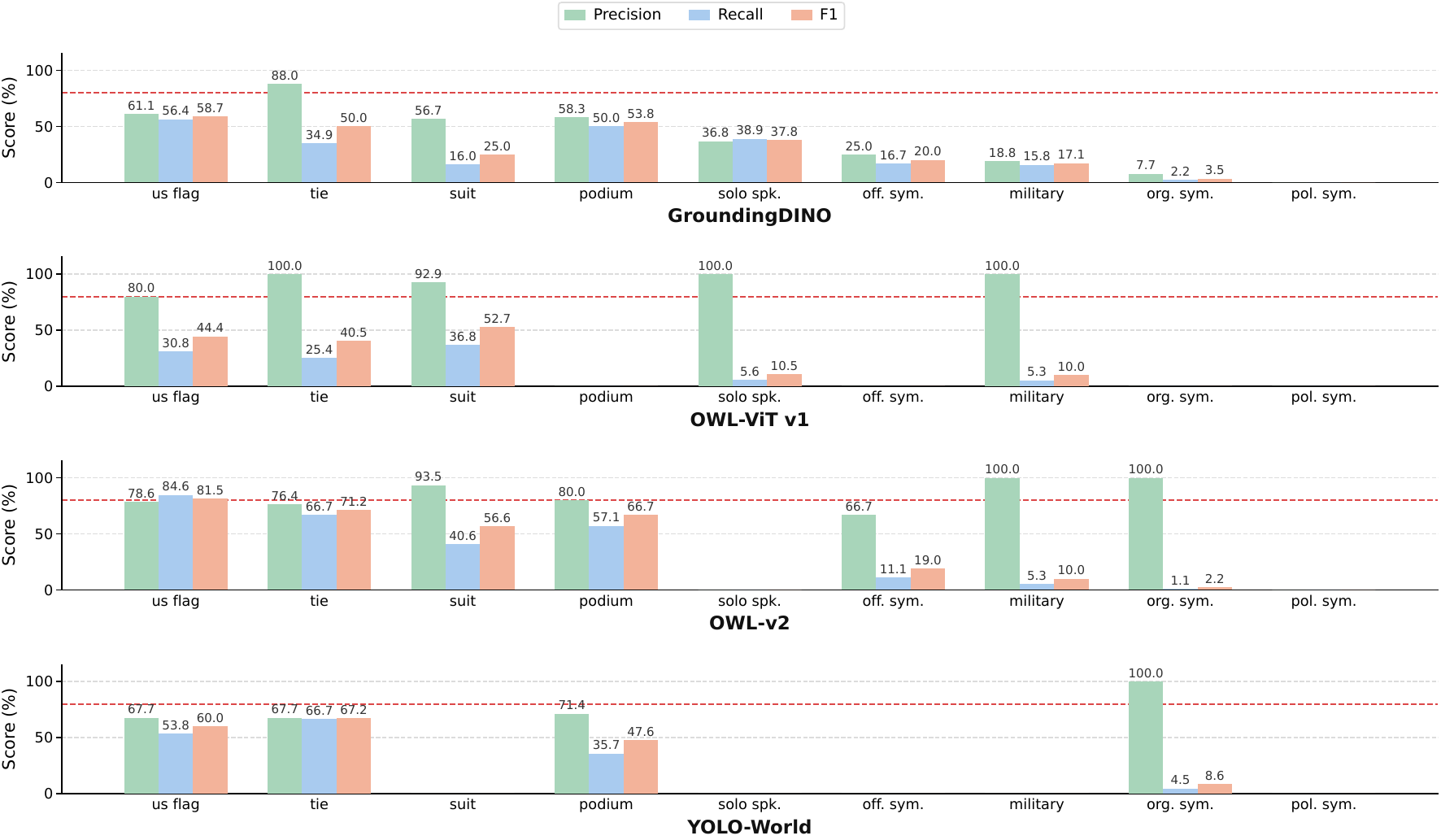}
  \vspace{-8pt}
    \caption{Performance breakdown across the nine target categories for each evaluated baseline model. Bar heights represent Precision, Recall, and F1 score. The red dashed line marks the 80\% precision threshold. The results illustrate a clear performance gradient, where models achieve high precision on concrete visual indicators but fail to reliably detect abstract sociological symbols.}
  \label{fig:performance_bar}
  \vspace{-15pt}
\end{figure*}

\subsection{Model Performance Evaluation and Calibration Impact}
\label{sec:overall_results}
Table~\ref{tab:main_results} presents the micro-averaged performance of the four models before and after semantic calibration. Our pipeline substantially improves precision across all architectures. The calibrated OWL-v2 reaches 80.2\% precision, a 43.3 percentage point gain over its zero-shot baseline. While stricter LLM filtering inevitably reduces recall, this conservative trade-off is desirable for computational social science applications, where measurement noise can propagate into downstream substantive findings. Overall F1 still improves for three of the four models, indicating that semantic false positives are removed without severely compromising detection capacity.

Figure~\ref{fig:performance_bar} breaks performance down by category, revealing a clear gradient aligned with the theoretical spectrum of visual abstraction. Concrete indicators with stable morphological structures, including \textit{U.S. flag}, \textit{tie}, and \textit{suit}, consistently approach or exceed the 80\% precision threshold under the OWL family. In contrast, abstract categories such as \textit{organization symbol} and \textit{politician symbol} yield near-zero recall and poor precision across all four detectors. This divergence empirically delineates the operating boundary of current vision-language models. They reliably ground distinct visual geometries but struggle to recognize abstract sociological concepts in the absence of explicit contextual supervision.

\subsection{Applying the Audit Framework}
\label{sec:applying_audit}
Prior to full-corpus deployment, we apply the three-tier audit protocol to determine the analytical viability of each target category.

\paragraph{Tier 1: Empirical Profile.}
We evaluate the calibrated pipeline across all four detectors on the nine target categories, yielding the per-category precision and recall reported in Figure~\ref{fig:performance_bar}. The profile traces the concrete-to-abstract gradient defined in previous section: concrete categories with stable visual prototypes, such as \textit{U.S. flag}, \textit{podium}, and \textit{suit}, attain precision above $75\%$ together with non-trivial recall under at least one detector; categories that require relational or contextual reasoning, such as \textit{solo speaker} and \textit{military}, retain reasonable precision but collapse to single-digit recall; and abstract categories without fixed morphological form, including \textit{organization symbol} and \textit{politician symbol}, fail on both axes. 

Among the categories that remain analytically viable, error profiles further differentiate the permissible analyses. Under the best-performing configuration, \textit{U.S. flag} achieves both high precision (78.6\%) and high recall (84.6\%), capturing most true instances while avoiding false detections and supporting absolute frequency estimation. \textit{Podium} and \textit{suit} maintain high precision ($\geq 80\%$) but substantially lower recall (57.1\% and 40.6\%). This asymmetry makes them suitable for relative comparisons across groups, where systematic undercounting cancels out, but unsuitable for absolute prevalence claims. Table~\ref{tab:tier1} reports the detailed metrics with bootstrap 95\% confidence intervals over images. The interval for \textit{U.S.\ flag} is the narrowest (P $[67.6, 90.0]$, R $[72.2, 94.6]$), which matters because it is the only category used for absolute frequency correction; \textit{podium} and \textit{suit} carry wider intervals, but their error profiles enter only the decision to restrict them to relative comparisons, not any reported estimate.

\begin{table}[t]
\centering
\small
\setlength{\tabcolsep}{3pt}
\begin{tabular}{l c c c c p{2.5cm}}
\toprule
Category & $n$ & P (\%) & R (\%) & F1 (\%) & Supported Analysis \\
\midrule
\textit{U.S. flag} & 39 & 78.6 & 84.6 & 81.5 & Absolute frequency, relative comparisons \\
\textit{Podium}    & 14 & 80.0 & 57.1 & 66.7 & Relative comparisons only \\
\textit{Suit}      & 106 & 93.5 & 40.6 & 56.6 & Relative comparisons only \\
\bottomrule
\end{tabular}
\caption{Tier 1 empirical profile for the three categories selected for full-corpus analysis. Metrics are derived from the calibrated OWL-v2 pipeline (IoU $\geq 0.5$); $n$ is the number of annotated instances. The ``Supported Analysis'' column reflects the task-oriented assessment dictated by each category's specific error profile.}
\label{tab:tier1}
\end{table}

\paragraph{Tier 2: Model Selection and Category Inclusion.}
Our downstream analyses compare symbol prevalence across political subgroups and over time, requiring strict precision together with adequate recall to maintain statistical power. Among the four detectors, OWL-v2 is the only one that satisfies both requirements on the three candidate categories identified in Tier 1. We therefore retain \textit{U.S. flag}, \textit{podium}, and \textit{suit} under OWL-v2 for full-corpus deployment and exclude the remaining six categories.

\paragraph{Tier 3: Prevalence Calibration.}
The performance asymmetry across the retained categories bounds the claims we can support. For \textit{U.S.\ flag}, high recall ensures that raw detection counts closely approximate true prevalence, and the Tier 3 correction (Equation~\ref{eq:tier3}) is applied to obtain absolute frequency estimates; we therefore report both raw and adjusted prevalence. For \textit{suit} and \textit{podium}, the low-recall regime introduces a distinct problem: although the Tier~3 estimator can in principle correct for undercounting, the large $1/R$ multiplier severely amplifies sampling uncertainty. We therefore report only relative comparisons for these indicators, specifically odds ratios and proportion differences between subgroups. This strategy assumes a uniform miss rate across political affiliations, which is defensible because the vision model evaluates purely visual properties such as occlusion or image quality, without access to partisan metadata. The ratio of detection frequencies between groups therefore approximates the true frequency ratio despite absolute undercounting.



\subsection{Case Study Results}

\begin{figure}[!t]
  \centering
  \includegraphics[width= \linewidth]{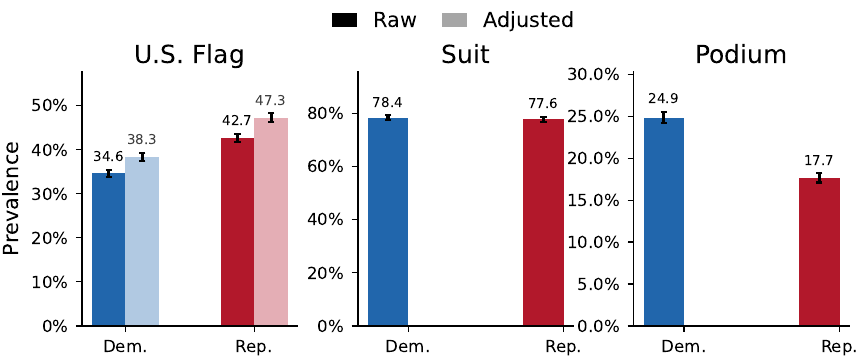}
    \caption{Detection prevalence across parties. For U.S.\ flag, bars show both raw rates and Tier~3 adjusted prevalence (Eq.~\ref{eq:tier3}), using image-level Precision and Recall estimated on the annotated sample; adjustment is not applied to \textit{suit} and \textit{podium} due to low recall (Table~\ref{tab:tier1}). Error bars show standard errors across legislators.}
\label{fig:prevalence}
\end{figure}

\subsubsection{Party Differences in Symbol Prevalence}
Figure~\ref{fig:prevalence} reports the prevalence of the three visual indicators across Democratic and Republican legislators. Republicans display the U.S.\ flag substantially more often than Democrats (42.7\% vs.\ 34.6\% of images). After Tier~3 correction (Eq.~\ref{eq:tier3}), the adjusted prevalence remains clearly higher for Republicans (47.3\% vs. 38.3\%), confirming that the gap is not an artifact of detection error. The image-level correction factor is $P/R = 1.11$ (bootstrap 95\% CI $[0.96, 1.32]$); the party comparison does not depend on it, since both groups share the same factor. For \textit{suit} and \textit{podium} we report only raw rates, since low recall (Table~\ref{tab:tier1}) precludes reliable absolute-frequency estimation. The raw rates already indicate near-identical suit usage across parties and substantially more podium appearances among Democrats.

\begin{table}[t]
\centering
\footnotesize
\setlength{\tabcolsep}{3.5pt}
\begin{tabular}{lccccc}
\toprule
\textbf{Party} & \textbf{Total} & \textbf{Flag} & \textbf{Rate} & \textbf{Adj. Cnt} & \textbf{Adj. Rate} \\
\midrule
Democrat   & 20,735 & 7,101 & 0.342 & 7,859.24 & 0.379 \\
Republican & 23,462 & 9,763 & 0.416 & 10,805.48 & 0.461 \\
\midrule
\multicolumn{6}{l}{\textbf{Association test}} \\
\multicolumn{6}{l}{$\chi^2(1)=252.75,\ p=6.54\times10^{-57},\ V=0.0756$} \\
\multicolumn{6}{l}{\textbf{Odds ratio (R vs.\ D)}: 1.368} \\
\bottomrule
\end{tabular}
\caption{Party differences in U.S.\ flag display. Republicans are 1.37 times as likely to display a U.S.\ flag.}
\label{tab:rq1_party_flag}
\end{table}
Table~\ref{tab:rq1_party_flag} formalizes the flag comparison through a $\chi^2$ test of independence. The association between party and flag display is highly significant ($\chi^2(1) = 252.75$, $p < 10^{-56}$), with an odds ratio of 1.37. Partisan differences in patriotic symbol display, long hypothesized in political communication research \citep{schill2012visual}, thus manifest at measurable scale in legislators' self-curated visual output.

\begin{figure}[!t]
  \centering
  \includegraphics[width= .8\linewidth]{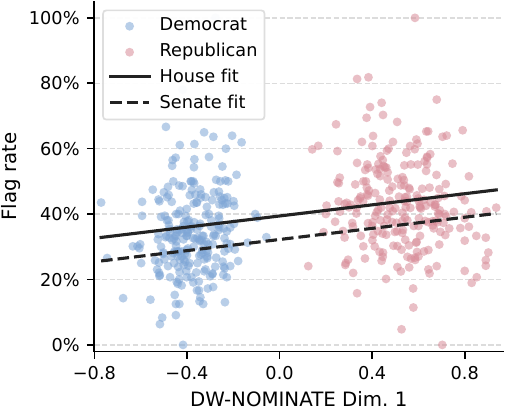}
    \caption{Ideology and U.S.\ flag display at the legislator level. DW-NOMINATE dimension 1 captures liberal--conservative positioning; separate OLS fits are shown for House and Senate. More conservative legislators display flags more frequently, with a consistent effect across chambers.}
\label{fig:ideology}
\vspace{-15pt}
\end{figure}

Party affiliation, however, is a coarse proxy for ideological position. Figure~\ref{fig:ideology} examines whether the effect is driven by the party boundary itself or by continuous ideological variation. Each point represents one legislator, plotted against DW-NOMINATE dimension 1. Separate OLS fits for House and Senate members reveal a consistent positive slope: more conservative legislators display U.S.\ flags more frequently, regardless of chamber. The effect is not a party step-function but a graded association with ideology.

\begin{table}[t]
\centering
\footnotesize
\setlength{\tabcolsep}{5pt}
\begin{tabular}{llll}
\toprule
\textbf{Comparison} & \textbf{Category} & \textbf{OR} & \textbf{95\% CI} \\
\midrule
\multicolumn{4}{l}{\textbf{Panel A: Republican vs.\ Democrat}} \\
\midrule
& Suit      & 0.934$^{*}$   & [0.892, 0.979] \\
& Podium    & 0.686$^{***}$ & [0.655, 0.718] \\
& U.S.\ Flag & 1.368$^{***}$ & [1.316, 1.422] \\
\midrule
\multicolumn{4}{l}{\textbf{Panel B: Senate vs.\ House}} \\
\midrule
& Suit      & 1.330$^{***}$ & [1.248, 1.418] \\
& Podium    & 1.020         & [0.960, 1.082] \\
& U.S.\ Flag & 0.737$^{***}$ & [0.700, 0.776] \\
\bottomrule
\end{tabular}

\vspace{2pt}
\begin{minipage}{0.95\linewidth}
\footnotesize
\textbf{Notes:} Odds ratios (OR) with 95\% CIs; OR $>1$ indicates the first group is more likely to display the category. Significance: $^{*}p<0.05$, $^{**}p<0.01$, $^{***}p<0.001$.
\end{minipage}
\caption{Relative comparisons in visual category display by party and chamber (RQ2). Republicans are less likely than Democrats to display suits and podiums but more likely to display U.S.\ flags. Senators are more likely than House members to display suits and less likely to display U.S.\ flags, with no significant chamber difference for podiums.}
\label{tab:rq2_odds_ratios}
\end{table}

Table~\ref{tab:rq2_odds_ratios} extends these comparisons to \textit{suit} and \textit{podium} through odds ratios with 95\% confidence intervals. Panel A shows that Republicans are significantly less likely than Democrats to display suits (OR $= 0.93$) and podiums (OR $= 0.69$) while more likely to display flags (OR $= 1.37$), suggesting that Democrats more frequently appear in formal indoor speaking settings whereas Republicans foreground patriotic symbols. Panel B compares chambers: senators are more likely to wear suits (OR $= 1.33$) and less likely to display flags (OR $= 0.74$), consistent with the Senate's more formal institutional culture and more heterogeneous state-level constituencies.

\begin{table*}[t]
\centering
\small
\setlength{\tabcolsep}{6pt}
\begin{tabular}{lcccccc}
\toprule
\textbf{Metric} & \textbf{Dem.\ Mean} & \textbf{Dem.\ Median} & \textbf{Rep.\ Mean} & \textbf{Rep.\ Median} & \textbf{Welch's $t$ ($p$)} & \textbf{Mann--Whitney $p$} \\
\midrule
Flag area ratio        & 0.0392 & 0.0323 & 0.0489 & 0.0423 & $-3.495$ ($5.20\times10^{-4}$) & $8.55\times10^{-6}$ \\
Flag center distance   & 0.4689 & 0.4686 & 0.4582 & 0.4558 & $1.833$ ($6.74\times10^{-2}$)  & $1.74\times10^{-2}$ \\
\bottomrule
\end{tabular}
\vspace{2pt}
\begin{tabular}{lrrrrrr}
\toprule
\textbf{Variable} & \textbf{Coef.} & \textbf{Std. Err.} & \textbf{$t$-statistic} & \textbf{$p$-value} & \textbf{CI Low} & \textbf{CI High} \\
\midrule
\multicolumn{7}{l}{\textbf{Panel A: OLS for flag area ratio}} \\
\midrule
Intercept      & 0.0367$^{***}$ & 0.0013 & 28.850 & $<10^{-16}$         & 0.0342 & 0.0392 \\
nominate\_dim1 & 0.0122$^{***}$ & 0.0026 & 4.700  & $2.61\times10^{-6}$ & 0.0071 & 0.0173 \\
is\_senate     & -0.0028        & 0.0029 & -0.940 & 0.347               & -0.0086 & 0.0030 \\
\midrule
\multicolumn{7}{l}{\textbf{Panel B: OLS for flag center distance}} \\
\midrule
Intercept      & 0.4623$^{***}$ & 0.0028 & 163.200 & $<10^{-16}$        & 0.4568 & 0.4679 \\
nominate\_dim1 & -0.0100$^{\dagger}$ & 0.0052 & -1.935 & 0.053          & -0.0202 & 0.0001 \\
is\_senate     & 0.0011         & 0.0064 & 0.177  & 0.860               & -0.0114 & 0.0136 \\
\bottomrule
\end{tabular}
\vspace{4pt}
\begin{minipage}{0.95\textwidth}
\footnotesize
\textbf{Notes:} Upper panel: distribution comparison by party at the legislator level ($N=470$ legislators; 239 Democrat, 231 Republican), where each legislator contributes the mean over their own flag detections. Lower panels: OLS models estimated on $N=29{,}178$ flag detections (12,174 Democrat, 17,004 Republican) nested within 470 legislators, with \texttt{nominate\_dim1} (DW-NOMINATE dim.\ 1; higher = more conservative) and \texttt{is\_senate} (Senate indicator). Standard errors are clustered within legislators. Significance: $^{\dagger}p<0.10$, $^{*}p<0.05$, $^{**}p<0.01$, $^{***}p<0.001$.
\end{minipage}
\caption{Spatial prominence of U.S.\ flags by party. Republicans' flags occupy a larger share of the image area; Democratic flags are slightly more centered but not statistically significant. Standard errors are clustered at the legislator level.}
\label{tab:flag_spatial_prominence}
\vspace{-15pt}
\end{table*}

\subsubsection{Spatial Prominence of National Symbols}
Beyond presence and absence, the pipeline enables fine-grained measurement of how visual symbols are displayed. Table~\ref{tab:flag_spatial_prominence} analyzes two continuous properties of detected U.S.\ flags: the \textit{area ratio} (fraction of image occupied by the flag mask) and the \textit{normalized center distance} (distance from the flag centroid to the image center, scaled to $[0, 1]$). Because detections are nested within legislators, comparisons are conducted at the legislator level and regression standard errors are clustered by legislator.

The upper panel reports distributional comparisons across parties. Republican flags occupy a larger image area on average (mean $0.0489$ vs.\ $0.0392$; Welch's $t = -3.50$, $p < 10^{-3}$), while Democratic flags are slightly more centered (mean center distance $0.4689$ vs.\ $0.4582$; $p = 0.067$). Both effects replicate under the non-parametric Mann--Whitney test ($p < 10^{-5}$ and $p = 0.017$).

The lower panels link these spatial properties to continuous ideology through OLS regression. Conservative ideology is associated with larger flag area ($\beta = 0.0122$, $p < 10^{-5}$) and, more tentatively, with less central flag placement ($\beta = -0.0100$, $p = 0.053$). Conservative legislators thus do not merely display flags more frequently; their flags are visually larger, with a weaker tendency toward peripheral placement. One plausible interpretation is that conservative legislators more often use flags as backdrop elements behind speakers or stages, whereas liberal legislators more often feature them as discrete central objects. This pattern is invisible to binary detection and demonstrates the analytical value of pixel-level spatial quantification.

\begin{figure}[!t]
  \centering
  \includegraphics[width= \linewidth]{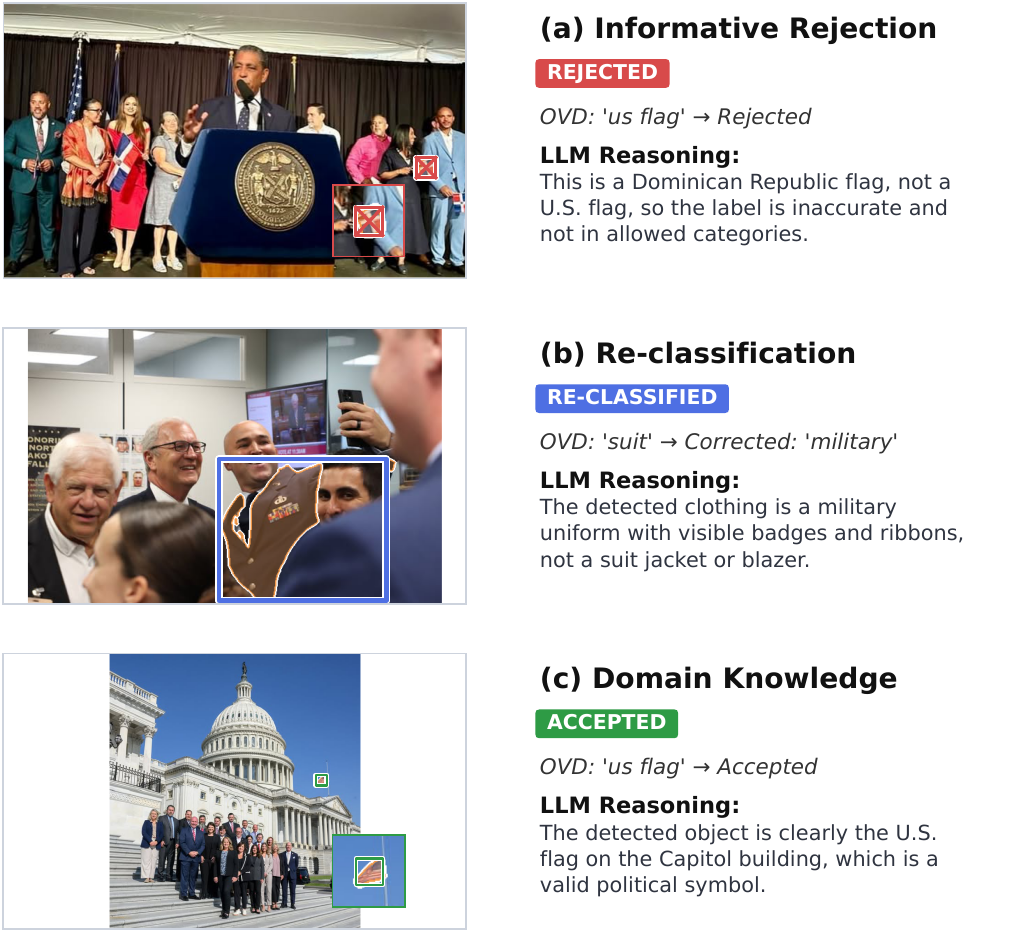}
\caption{Qualitative examples of LLM-as-Judge decisions. (a) A Dominican Republic flag is rejected as non-U.S.; (b) a military uniform is re-classified from `suit' (OWL-ViT~v1 example); (c) a U.S.\ flag on the Capitol is accepted. The natural-language reasoning externalizes domain knowledge, flag origin, uniform insignia, landmark identification, that would otherwise require trained annotators. Orange: SAM mask; box color: red=rejected, blue=re-classified, green=accepted.}
\label{fig:qualitative}
  \vspace{-15pt}
\end{figure}

\subsection{Qualitative Analysis of LLM Reasoning}
Beyond quantitative patterns, the natural-language reasoning chains produced by the LLM reveal a qualitative benefit relevant to practical deployment. Figure~\ref{fig:qualitative} shows three representative cases. In Case (a), an OVD detector confidently labels a flag as ``us flag'' (confidence $0.47$); the LLM rejects the detection, identifying the object as a Dominican Republic flag. In Case (b), a detection originally labeled ``suit'' is re-classified as ``military'' once the LLM recognizes the characteristic badges and ribbons of a military uniform. In Case (c), the LLM accepts a U.S.\ flag detection and grounds the decision in contextual evidence, identifying the Capitol behind the subjects.

These cases share a common property: the LLM externalizes domain knowledge that would otherwise reside with trained annotators. Distinguishing foreign flags from U.S.\ flags, identifying military insignia, and recognizing U.S.\ government landmarks are non-trivial visual competencies. By producing reasoning chains that state this knowledge explicitly, the pipeline shifts human-in-the-loop validation from a task requiring visual expertise to one requiring only logical evaluation of the stated rationale. A non-expert reviewer can assess whether ``this is a Dominican Republic flag'' is a well-reasoned decision without being able to identify every national flag on sight. This property lowers the expertise threshold for quality control, a practical benefit for research teams without specialized visual-communication expertise.

\section{Discussion and Conclusion}
We presented \textit{Perceive, Refine, Reason} (PRR), a calibrated pipeline that augments off-the-shelf open-vocabulary detection with SAM-based spatial quantification and LLM-based semantic arbitration. The accompanying three-tier audit framework turns imperfect detections into measurements with documented bias bounds. Applied to 103{,}920 Facebook images posted by U.S.\ legislators during the 2024 election cycle, the framework supported party-level prevalence comparisons, regression of visual indicators on continuous DW-NOMINATE ideology scores, and pixel-level analysis of spatial prominence. Beyond confirming known partisan asymmetries in flag display, the spatial analysis revealed a pattern invisible to binary detection: more conservative legislators display U.S.\ flags as larger elements, with a weaker tendency toward peripheral placement.

\paragraph{Implications for computational social science.}
The framework offers a ready-to-use toolkit for measuring fine-grained visual indicators at scale. Categories are defined in natural language, measurements are spatially grounded, and the audit framework makes the boundaries of reliable inference explicit. The design is model-agnostic: the SAM and LLM calibration backbone can be paired with any OVD front-end as new detectors emerge. The natural-language reasoning chains further externalize domain knowledge that would otherwise reside with trained annotators, lowering the expertise threshold for human-in-the-loop validation.

\paragraph{Limitations.}
Several limitations bound our conclusions. The gold-standard test set comprises 105 images (220 annotated instances); because the audit protocol is modular, a larger test set tightens the reported intervals without re-running detection, making annotation budget an explicit parameter of the framework. The case study covers one platform, one country, and one election cycle, and generalization requires re-profiling on representative test sets. The LLM arbitration layer inherits biases from its training data, and the audit framework documents aggregate error rates but not differential errors across demographic subgroups. Finally, the ideology-imagery associations are correlational, and we cannot identify whether individual image decisions reflect legislators, communications staff, or platform curation.

\paragraph{Future work.}
A richer audit framework could profile differential error rates across demographic subgroups to detect representational biases. Empirically, the pipeline extends naturally to other platforms such as Instagram and TikTok, other substantive domains such as public health and protest mobilization, and longitudinal corpora that capture how visual rhetoric evolves over time.

\section{Ethical Considerations}
This work analyzes publicly posted images from the official Facebook accounts of elected federal officials acting in their professional capacity, obtained through documented public APIs. Because legislators' official social media output is a form of public political communication, the privacy expectation differs from that of ordinary user content. We nonetheless restrict our analyses to visual indicators directly relevant to the stated research questions.

The pipeline carries potential dual-use risks: the same components that enable aggregate measurement could be repurposed to surveil non-public actors or profile individuals in harmful ways. We mitigate these concerns by framing the methodology around aggregate measurement rather than individual classification, tying every analytical claim to the calibrated reliability of each category, and releasing only the pipeline, gold-standard annotations, and aggregate statistics rather than the image corpus itself. A further concern is that the LLM arbitration layer may produce differential reliability across demographic groups underrepresented in pre-training data. Researchers deploying this pipeline in fairness-consequential domains should profile per-subgroup error rates as part of Tier~1 calibration.

\bibliographystyle{ACM-Reference-Format}
\bibliography{WSDM/main}

\end{document}